\documentclass[a4paper, 10pt, conference]{ieeeconf}      
\usepackage{FG2025}

\FGfinalcopy 

\IEEEoverridecommandlockouts                              
\def\FGPaperID{008} 

\title{\LARGE \bf
My Emotion on your face: The use of Facial Keypoint Detection to preserve Emotions in Latent Space Editing
}

\author{\parbox{16cm}{\centering
    {\large Jingrui He and  A. Stephen McGough}\\
    {\normalsize
     School of Computing, Newcastle University, UK. \{j.he23, stephen.mcgough\}@newcastle.ac.uk\\
    }}
}
\usepackage{graphicx}
\usepackage{amsmath}
\usepackage{amssymb}
\usepackage{booktabs}
\usepackage{makecell}
\usepackage{multirow}
\usepackage{array}
\usepackage{tabularx}
\newcolumntype{Y}{>{\centering\arraybackslash}X}
\newcommand{\figref}[1]{Fig.~\ref{#1}}
\usepackage{stfloats}

\newcommand{\etal}{{\em et al.}}

\usepackage{fancyhdr}
\begin{document}

\ifFGfinal
\thispagestyle{empty}
\pagestyle{empty}
\else
\author{Anonymous FG2025 submission\\ Paper ID \FGPaperID \\}
\pagestyle{plain}
\fi
\maketitle

\thispagestyle{fancy}

\begin{abstract}
Generative Adversarial Network approaches such as 
StyleGAN and StyleGAN2 provide two key benefits: the ability to generate photo-realistic face images and possessing a semantically structured latent space from which these images are created. 
Many approaches have emerged for editing images derived from vectors in the latent space of a pre-trained StyleGAN and StyleGAN2 models by identifying semantically meaningful directions (e.g., gender, age or hair color) in the latent space.
By moving the vector in a specific direction, the ideal result would only change the target feature while preserving all the other features. 
Providing an ideal data augmentation approach for gesture research as it could be used to generate numerous image variations whilst keeping the facial gestures and expressions intact.
However, the entanglement issue, where changing one feature inevitably affects other features, impacts the ability to preserve facial gestures and expressions. 
To address the entanglement issue, we propose the use of an addition to the loss function of a Facial Keypoint Detection model to restrict changes to the facial gesture.
Our approach builds on top of an existing model and simply adds the proposed Human Face Landmark Detection (HFLD) loss, provided by a pre-trained Facial Keypoint Detection model, to the original loss function.
We quantitatively and qualitatively evaluate the existing and our extended model, showing the effectiveness of our proposed loss term in addressing the entanglement issue and maintaining the facial gesture and expression.
Our approach achieved up to 49\% reduction in the change of emotion in our experiments.
Moreover, we show the benefit of our approach by comparing with state-of-the-art models. 
By increasing the ability to preserve the facial gesture and expression during facial transformation, we present a way to create human face images with fixed expression but different appearances, making it a reliable data augmentation approach for Facial Gesture and Expression research.

\end{abstract}

\section{INTRODUCTION}
\label{sec:intro}
Development of Generative Adversarial Networks have revolutionized our ability to generate synthetic images. 
This is especially prevalent for the generation of photo-realistic human face images using models such as StyleGAN~\cite{karras2019style} and StyleGAN2~\cite{karras2020analyzing}.
These models derive images from a vector in a latent space -- often referred to as a latent code. 
With these latent spaces demonstrating semantic relations -- a latent code can be transformed in a specific way to obtain a desired feature change, e.g., we can change someone's hair type by moving in the direction of ``wavy hair''~\cite{huang2023adaptive}.

This would appear to be ideal for Facial Gesture research as it provides an endless supply of synthetic face images. 
However, there is no clear mechanism to create images with a pre-defined facial gesture. 
Selecting random latent codes would provide images, but with no way of locking down the facial gesture produced. 
Existing works~\cite{harkonen2020ganspace,huang2023adaptive,shen2020interfacegan,shen2021closed} provide mechanisms for changing one feature (e.g., hair color) of the new image, however, due to feature entanglement it is not possible to change one feature without affecting others, negating their use without manual labeling. 
Instead we need an approach which preserves the facial gesture as part of the transformation process allowing labels to be automatically mapped to the newly generated images.

Approaches for facial transformations -- commonly referred to as Face Editing -- aim to change one feature given a synthetic image (from a known latent code) or even starting from a real image.
There are two main approaches to this.
1) An end-to-end transformation structure -- referred to image space editing~\cite{zhuang2021enjoy}. 
Most image space editing methods use an auto-encoder style structure to perform cross-domain transformations. 
For example, in the context of facial images, the source domain could be ``female'', and the destination domain could be ``male''. 
After training, given a female image, the model would transform this into a male version of the original. 
However, this method is often associated with excessive computing cost as a generative model needs training from scratch for each desired transformation, restricting its use for generating large numbers of images.
In addition, unless carefully trained the model would not guarantee to preserve gestures.
2) Latent space editing, which uses a pre-trained GAN and attempts to analyse the latent space of the GAN to identify meaningful semantic directions within the space -- to change just one feature. 
For example, moving in a particular direction will make the face ``older''.
StyleGAN~\cite{karras2019style} and improved version, StyleGAN2~\cite{karras2020analyzing}, have demonstrated both an ability to generate photo-realistic images of faces but also demonstrated a good semantic structure to their latent space making them an ideal choice for many researchers.
However, it has not been possible, to date, to disentangle different features when editing an image -- changing the age of a face will often change the expression at the same time. 
Reducing the applicability of this approach for generating many different faces which possess the same facial gesture.

\begin{figure*}[!t]
  \centering
   \includegraphics[width=0.85\linewidth]{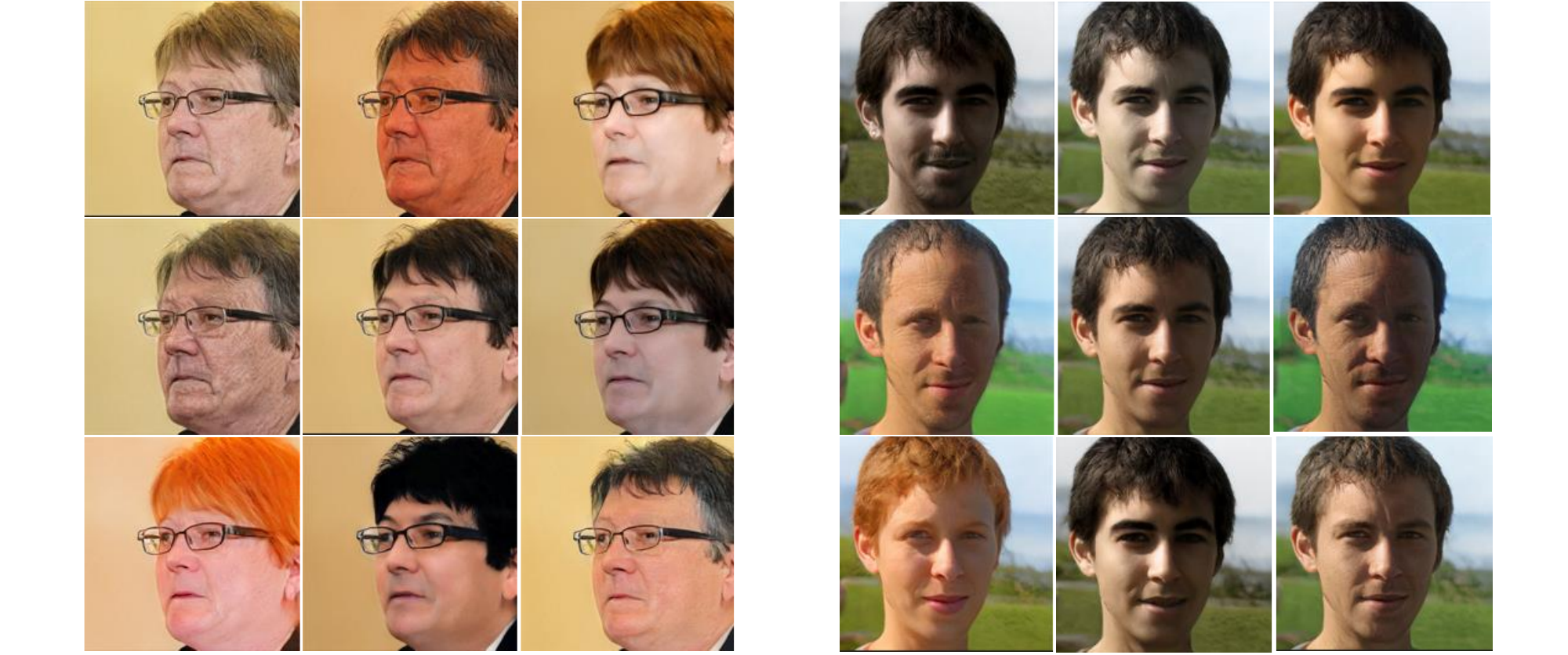}
   \caption{We present the variations offered by our approach. The middle image is the original one, all the surrounding images are edited versions. We illustrate that all image have the different appearances but same facial gesture and expression.}
   \label{showcase}
\end{figure*}

In order to determine meaningful semantic directions, latent space editing can be divided into supervised methods and unsupervised methods.
GANSpace~\cite{harkonen2020ganspace} used an unsupervised method to analyse the latent space of a pre-trained StyleGAN. 
Using Principle component analysis (PCA), each PCA direction presents a semantic direction in latent space, meaning that moving in that directions can achieve meaningful feature changes in the generated images.
Instead of using an unsupervised approach, InterFaceGAN~\cite{shen2020interfacegan} used an additional classifier to get the attribute-related positive and negative samples and then trained an SVM to find a boundary, which can be used to edit the latent code. 
Enjoy your Editing~\cite{zhuang2021enjoy} used a facial feature classification model as a guide to find a set of directions to achieve latent space editing.
StyleFlow~\cite{abdal2021styleflow} resampled the latent code with the normalizing flow guided by the target attributes.
AdaTrans~\cite{huang2023adaptive} divides the non-linear transformation process into several steps.
Similarly, SDFlow~\cite{li2024semantic} is a flow-based model with a semantic encoder to demonstrate the
effectiveness in terms of editing accuracy and disentanglement.
Unfortunately, none of these approaches have solved the entanglement problem. 

Thus, we need a more precise and disentangled approach. 
Ideally, changing one attribute should not affect other attributes at all. For example, if the requirement is to change the hair color, then only the hair color should be changed. 
All other attributes, e.g.,  expression, age, and gender, should be unaffected. 
Although some works can reduce this issue to some extent, the performance is degraded when the original facial image is in an extreme state, e.g., extreme head orientation, or slightly exaggerated expression.

In this paper, we propose using facial keypoint detection (eyes, nose, lips etc.) to constrain the process of identifying feature movements in the latent space. 
By performing this constraint we minimize the change in the location of these keypoints thus maximizing the consistency of gesture between images when modifying a particular feature.
We exemplify this by adding a term to the loss function of  Enjoy your editing~\cite{zhuang2021enjoy} for Human Face Landmark Detection (HFLD). 
We demonstrate this in \figref{showcase}.
Though we expect similar results would be obtained for other latent space editing approaches.

The contribution of our work are: 
\begin{itemize}
    \item We analyse the entanglement issue in the baseline model (Enjoy Your Editing) from both a quantitive and qualitative perspective. 
    Highlighting those features which are most entangled. Section\ref{sec:analysis to the baseline model}.
    \item We propose a novel loss term, HFDL loss, and apply this to the Enjoy Your Editing model. Section \ref{sec:formatting}.
    \item We demonstrate the effectiveness of our proposed loss term, showing that it can significantly reduce the entanglement issues as well as preserve facial gestures.
    In addition our method outperforms the state-of-the-art methods in terms of disentanglement. Section \ref{sec:Experiments}.
\end{itemize}


\section{Related Work}
\label{sec:backg}
\subsection{Generative Adversarial Networks (GANs)}
Goodfellow \etal~\cite{goodfellow2014generative} proposed Generative Adversarial Networks (GANs), which trains two models (Generator and Discriminator) simultaneously in an adversarial set up. 
During training, the Generator learns to map a noise vector, referred to as latent code (in the latent space), to a high-quality image in the image space with the aim of preventing the discriminator from being able to distinguish the difference between synthetic and real images.
DCGAN~\cite{radford2015unsupervised} further improve the quality of the synthetic images by integrating Convolutional Neural Networks (CNNs) into GANs model -- as opposed to all fully connected layers. StyleGAN~\cite{karras2019style} further improved the performance of GANs.
Different from traditional GANs (e.g., the original GAN and DCGAN), StyleGAN maps the traditional latent space to a more disentangled space, called $w$-space, and inject multiple latent codes at different layers of generator, using adaptive instance normalization (AdaIN)~\cite{8237429}. 
The structure enables finer-grained control over visual features, thus producing photorealistic, high resolution images.
StyleGAN2~\cite{karras2020analyzing} went further by the removal of the artifact in the image referred to as the ``blob" issue, producing higher quality synthetic images.

\subsection{Latent Space Editing on Human Face Images}
StyleGAN/2 not only show a significant advantage in quality of synthetic image generation, but also present a highly disentangled latent space.
Many approaches have been proposed to achieve human face editing via latent space editing given a pre-trained StyleGAN/2 model~\cite{abdal2021styleflow,alaluf2021only,harkonen2020ganspace,huang2023adaptive,li2024semantic,parihar2022everything,shen2020interfacegan,shen2021closed,zhuang2021enjoy}. 
Specifically, latent space editing focuses on finding a way to move a latent code in a defined manner to enact a semantic change (e.g., gender or age) reflected in the generated image.


GANSpace~\cite{harkonen2020ganspace} applies PCA analysis on the $w$-space to discover directions which control different human face features (e.g.,  gender, age or smiling).
Allowing a latent code to be moved linearly in one of these feature directions to obtain a edited human face image where, ideally, the only change would be that of the chosen feature. 
Furthermore, they map to \(w^+\) space allowing them to inject different latent codes to different layers of the generator to achieve  precise control of the image editing.

A similar linear approach is presented by Shen \etal~\cite{shen2020interfacegan}.
They first use a facial feature classifier to identify the most positive and negative samples of one feature defined in the context of human face.
They then use an SVM to decide the binary boundary of this feature such as gender or age. 
In this way, they get the semantic editing directions in latent space and move the latent code to achieve latent space editing. 

Enjoy your Editing~\cite{zhuang2021enjoy} use an additional feature regressor to guide a linear model to solve the human face editing problem. 
The model takes in a latent code then outputs the transformed one according to the target feature, which is then passed to the generator to obtain the edited face image.

For non-linear editing, StyleFlow~\cite{abdal2021styleflow} used normalizing flows to map the latent space to a more disentangled space conditioned on the target facial feature. 
AdaTrans~\cite{huang2023adaptive} used a learnable network to non-linearly predict the best shift for an original latent code given a target feature for image editing.

However, the entanglement issue is still unsolved in both linear and non-linear settings.
This removes the ability to guarantee facial gestures are preserved.
Ideally, to augment a synthetic dataset for gesture research, the expression, head orientation and direction of gaze should be unaffected when modifying a feature of the face if we do not wish to manually label the new images. 

\subsection{Facial Expression Generation}
GANs show rich semantic information in latent space, which makes it a useful tool to generate images of human faces with different expressions.
ExprGAN~\cite{ding2018exprgan} uses GAN framework with an additional expression controller module which can produce a expressive code. This code is then used as a condition during image generation process to achieve continuous adjustment for expression editing.
StarGAN~\cite{choi2018stargan}, a conditional GAN model, uses a single generator and discriminator to handle multiple domain image-to-image translation simultaneously. 
With the help of expression condition within a discrete number of expressions, StarGAN can generate human face image with the target expression while maintaining other features as close as possible to the original image.
GANimation~\cite{pumarola2020ganimation} integrates attention mechanism to the GAN model, along with action units (AUs) as the expression labeling, generating a wide range of facial expressions with varying intensities. 
StarGAN v2~\cite{choi2020stargan}, built on top of StarGAN, uses the concept of style representation in a specific domain to produce diverse editing results given one image and a target domain instead of deterministic result in the original StarGAN.
GANmut~\cite{d2021ganmut} proposed a training scheme of conditional GANs.
By jointly optimizing the GAN model and the conditional latent space, the model can produce diverse and complex expressions using only basic expression labels.

However, existing approaches are either limited by a discrete set of labeled and basic expressions or lack evidence of performance under challenging conditions (e.g., extreme head orientations or random gaze directions).

In this paper, we propose a new loss addition (e.g., HFLD loss) to a baseline model, presenting the benefits in terms of preserving the facial gesture and expression.
We utilize the diverse image generation ability of SyleGAN2 model where the expression produced by the model is complex and closer to real world situation.
By fixing the expressions and gestures but generating different appearances of the original image, we provide an effective way to produce a large number of diverse human face images with a same realistic expression.


\section{Method}
\label{sec:formatting}
\subsection{Overview}
We use Enjoy your Editing~\cite{zhuang2021enjoy} as the baseline model.
The baseline aims to find a Matrix $T$ which contains semantic latent space directions to manipulate the generated images.
Built on top of their original training pipeline, we employ a pre-trained keypoint detection model to provide an additional HFLD loss to guide the training of the matrix $T$.
By restricting the difference of face landmark between edited image \(x^{\prime}\) and the original image \(x\), we are able to maximize the retention of 
facial expressions and gestures.

\begin{figure*}[htbp]
  \centering
   \includegraphics[width=1\linewidth]{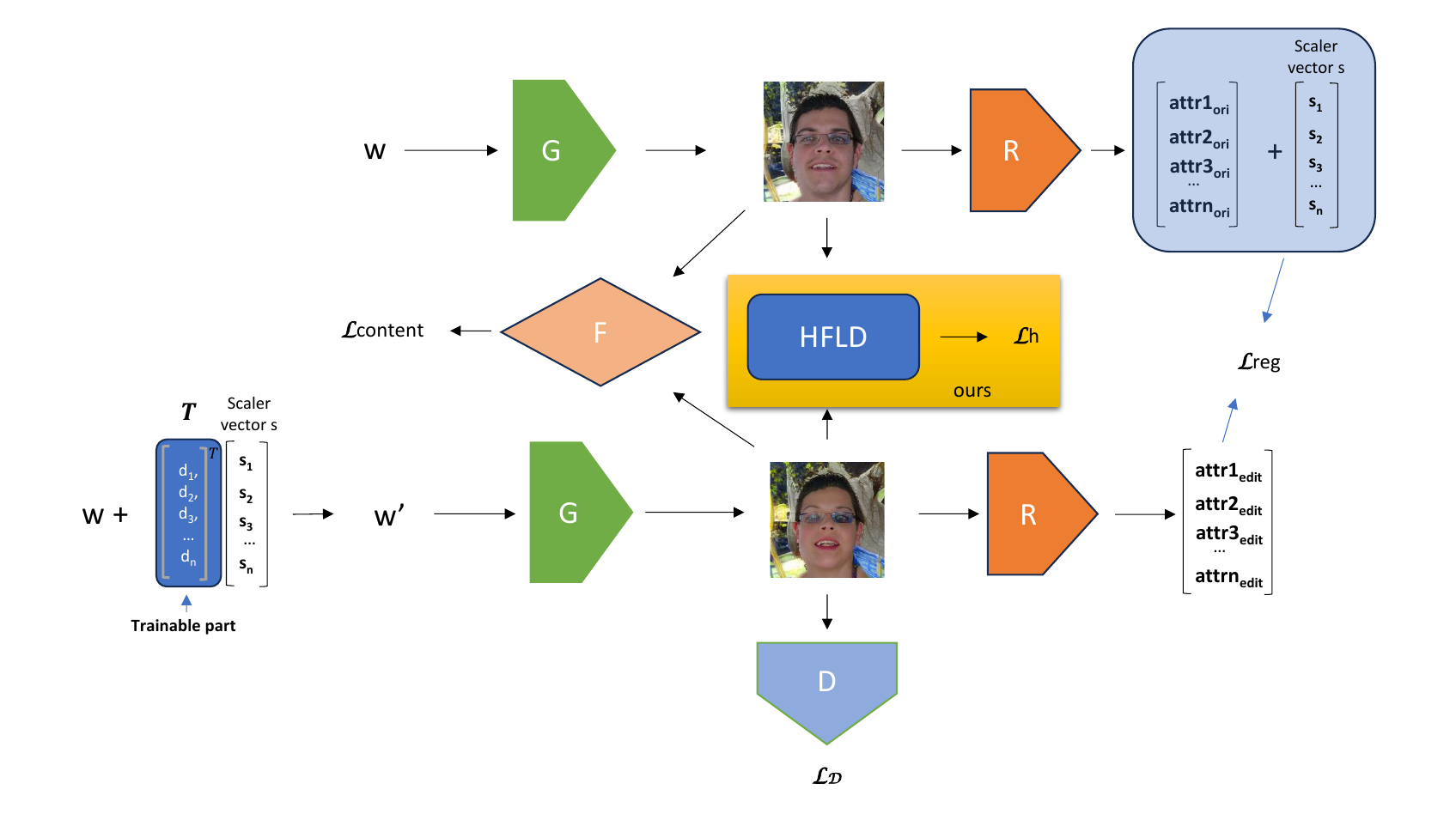}
   \caption{The overview of the training pipeline. The baseline approach (Enjoy your Editing) first sampled the scaler value $s$ for facial features, 40 elements in total. Given a scaler and original code $ w $, the trainable matrix $T$ is responsible for generating a editing shift with the degree control by the scalar vector $s$. The $ w^\prime $, obtained by $w + Ts$, is further passed through the Generator $G$ to get the edited image. The baseline method contains $\mathcal{L}_{reg}, \mathcal{L}_{D},\mathcal{L}_{content}$ calculated by pre-trained Regressor $R$, Discriminator $D$ and Feature extractor $F$ (Perceptual model-VGG19), respectively. Our proposed loss term $\mathcal{L}_{h}$ is calculated by the pre-trained HFLD model presented in yellow box. All the losses listed in the Figure are used for training of the matrix $T$.}
   \label{Overview}
\end{figure*}

\figref{Overview} illustrates the overall training pipeline. 
We use Enjoy your Editing~\cite{zhuang2021enjoy} as baseline and add HFLD guidance (see yellow box in \figref{Overview}) to their original training scheme.
Pre-trained StyleGAN2 generator, discriminator, perceptual model and the facial feature regressor are used as in the baseline. 
During training, the original image $x = G(w)$, where $w$ is the latent code from the latent space of StyleGAN2, is fed into the regressor $R$ to get the original feature vector $A_{ori} = \{a_1,a_2,...,a_n\}$. 
Then given a scaler vector $s = \{s_1,s_2,...,s_n\}$ drawn from a distribution $\mathcal{D}_s$ uniformed in $[-1, 1]^N$, the offset for \(w\) is calculated by the trainable matrix $T$ with the guidance of the scaler vector $s$. 
This offset is added to \(w\) to get the final result image $x^{\prime}$ = $G(w + Ts)$. 
Ideally, the predicted facial feature value of transformed image $A^{\prime} = R(x^{\prime})$ should approach the ground truth feature value $A_{gt} = A_{ori} + s$ as close as possible.

To preserve the image quality and increase the realism of the transformed image, the baseline uses a perceptual model and a discriminator to guide the training of the matrix.

Our proposed HFLD loss is added to their final loss and used in their original training process to preserve the facial gesture and expression.

\subsection{Architecture}

We adopt the models from Zhung \etal~\cite{zhuang2021enjoy} along with our HFLD model to provide losses for training.

\subsubsection{Baseline model}

\textbf{Trainable latent space directions matrix.}
As we use Zhuang \etal's \cite{zhuang2021enjoy}  ``global'' method for image transformation, the trainable part in the model is the latent space directions matrix $T = \{d_1,d_2,...,d_n\}$, which contains semantic meaningful directions (e.g, gender, age or hair color).
A scaler feature vector $s = \{s_1,s_2,...,s_n\}$ is used to control the changing degree of corresponding features in $T$.
As shown in (\ref{eq:important}), the final edited code is obtained by adding a shift derived from $Ts$ to the original code $w$.
\begin{equation}
  w^{\prime} = w + Ts
  \label{eq:important}
\end{equation}

In addition, as the experiments in Zhuang \etal~\cite{zhuang2021enjoy} show the editing in the $ w^+$ space performs better than the $w$ space, we use the same set up for designing the matrix, training it to get the directions in $w^+$ space. 

\textbf{Human face feature Regressor.} The feature regressor $R$ is a pre-trained model that is used to predict the facial feature values given an image. The output is a 40 dimensional vector where each element in the vector is a specific facial feature value (e.g., Male, Young or Smiling).

\textbf{Perceptual model-VGG19.}
The perceptual model $F$ is a pre-trained VGG19 model used to extract the features of an image. Unlike the pixel loss which can only determine the difference on the pixel level. The perceptual loss calculates the difference between the extracted features predicted by the VGG19 model, which can reveal the high-level feature similarity of compared images.

\subsubsection{HFLD model}
\begin{figure}[t]
  \centering
   \includegraphics[width=0.4\linewidth]{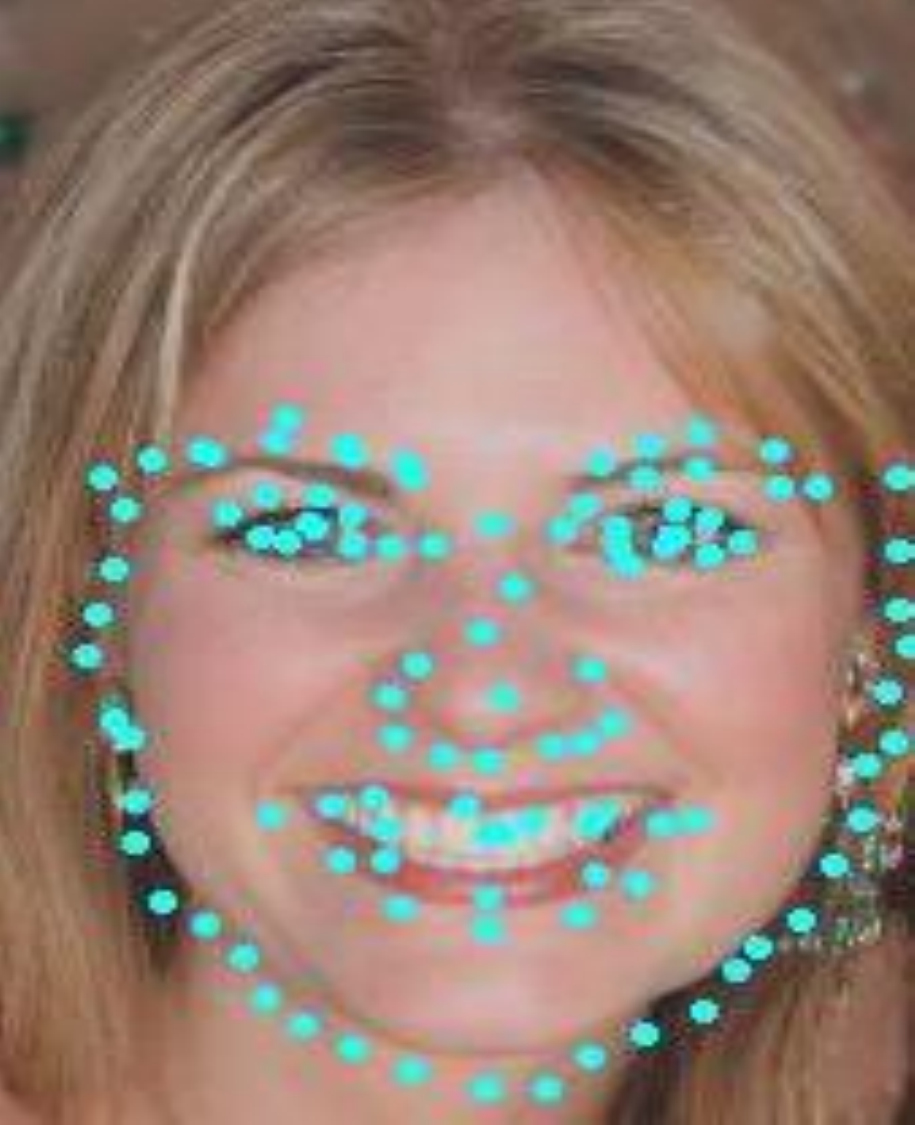}

   \caption{The prediction result using HFLD model. Given a facial image, the prediction is 106 landmark points.}
   \label{fig:HFLD RESULT}
\end{figure}

We apply our proposed HFLD model to guide the training. \figref{fig:HFLD RESULT} shows the result predicted by the model. The pre-trained model outputs 106 landmark points of the face area given a facial image. Our motivation is to preserve the facial gestures as much as possible. By restricting changes to landmarks, the transformed image is expected to retain gestures such as head orientation and expression. 

\subsection{Loss Function}
The trainable matrix $T$ is guided by regressor loss, content loss/perceptual loss, Discriminator loss and HFLD loss, where the proposed HFLD loss is specifically used to improve the disentangled performance and keep the facial gestures and expressions.

\subsubsection{Baseline loss}
The baseline loss has three parts: regressor loss, content loss and discriminator loss.

The regressor loss guides the matrix $T$ to perform image transformation by the control of a given scaler feature vector $s$. 
Thus the loss function can be interpreted as
\begin{equation*}
\mathcal{L}_{reg} = \mathbb{E}_{w \sim \mathcal{W},s \sim \mathcal{D}_{s}} [A^{\prime}\log A_{gt} - (1-A^{\prime})\log (1-A_{gt})].
  \label{eq:regressor loss}
\end{equation*}
$A^{\prime} = R(x^{\prime})$ is the predicted facial feature value of transformed image while $A_{gt}$ stands for the ground truth value. Since the latent code $w$ is sampled from $\mathcal{W}$ space, and the scaler feature vector $s$ is sampled from the distribution $\mathcal{D}_s$, the distribution of $A_{gt}$, which is derived from $A_{gt} = R(G(w)) +s$, is dependent on $\mathcal{W}$ and $\mathcal{D}_s$.

The Discriminator loss is the adversarial loss typically used in GAN training. In the baseline model, it is used to keep images realistic during facial transformation training. We use $\mathcal{W}^{\prime}|w$ to clarify that the predicted $w^\prime$  sampled from distribution $\mathcal{W}^{\prime}$ is conditioned on the original code $w$
\begin{equation*}
\mathcal{L}_{D} = \mathbb{E}_{w^{\prime} \sim \mathcal{W}^{\prime}|w} [ 
 \log (1- D(G(w^\prime)))  ].
  \label{eq:D loss}
\end{equation*}

The content loss used in the baseline involves using a pre-trained perceptual model-VGG19. By calculating the difference of extracted features between two images, the high-level similarity is measured. Finally, the loss is shown as follows 
\begin{equation*}
\mathcal{L}_{c} = \mathbb{E}_{w \sim \mathcal{W},w^{\prime} \sim \mathcal{W}^{\prime}|w} \sum_i || F_i(G(w)) - F_i(G(w^\prime)) ||_2 
  \label{eq:content loss}
\end{equation*}
where $F_i(\cdot)$ denotes the extracted feature maps at layer $i$ of the pre-trained perceptual model $F$ (i.e., perceptual model-VGG19).
\subsubsection{HFLD loss}

In order to maintain the gestures of the original facial image, we use a HFLD model to restrict undesired changes during transformation. 
Given an original image \(x\) and a edited image \(x^{\prime}\), the HFLD model calculates landmark points via $H(x)$ and $H(x^{\prime})$, respectively. 
We use the absolute difference between the two feature vectors extracted by HFLD model to provide the loss
\begin{equation*}
\mathcal{L}_{h} = \frac{1}{N} \sum_i^n (| H_i(x) - H_i(x^{\prime}) |)
  \label{eq:hfld loss}
\end{equation*}
where $N$ represents the dimension of the feature vector, in this case, 212 (106 landmark points in total).

Thus, the final loss is
\begin{equation*}
\mathcal{L} = \lambda_1 \mathcal{L}_{reg} + \lambda_2 \mathcal{L}_{c} + \lambda_3 \mathcal{L}_{D} + \lambda_4 \mathcal{L}_{h} 
  \label{eq:total loss}
\end{equation*}
where $\lambda_1, \lambda_2, \lambda_3, \lambda_4$ are hyperparameters to control the importance of each loss.


\section{Experiments}
\label{sec:Experiments}
\subsection{Implementation Details}
We used a pre-trained styleGAN2 model trained on the FFHQ~\cite{karras2019style} dataset. 
We used the pre-trained PyTorch weights of the StyleGAN2 model at a resolution of $256\times256$ to align with the baseline model and enable fair comparison. 
To make comparisons with other state-of-the-art models, we used official config-f weights provided by Nvidia~\cite{karras2019style} for the StyleGAN2 model. 
The VGG19 model is trained on the ImageNet dataset~\cite{russakovsky2015imagenet}. 
The regressor $R$ is a resnet50-based model where the last layer has been modified to a dense layer to meet the requirement of outputting the predicted feature value. 
In this case, the dimension of the last layer is 40, which is consistent with the number of labeled binary classes in the CelebA dataset~\cite{liu2018large}. For the HFLD model, we used resnet-18 as a base model and replaced the last layer with a 212-dim dense layer. 
The model takes $256\times256$ resolution images and outputs the predicted landmark points.
As for the matrix $T$, we trained for 50,000 iterations with a batch size of 4. The learning rate is set to $10^{-4}$.
We chose the same $ \lambda_1, \lambda_2, \lambda_3 $ values as the baseline and tuned $\lambda_4$ with the best performance for our HFLD loss.
We set $ \lambda_1, \lambda_2, \lambda_3, $ and $\lambda_4$ to 10, 0.05, 0.05, and $0.5 \times 10^{6}$ respectively.


\subsection{Quantitative analysis of the baseline and our model}
\label{sec:analysis to the baseline model}
To quantitatively compare with the baseline model, we first sampled 1,000 synthetic images and used the baseline approach and our model to perform transformation according to the target feature. 
For the baseline and our models, we performed transformation on target features at two different scales, where the target feature values are set to zero (non-exist) and one (must-exist). 

After gaining two completely opposite sets of images, we calculated the mean difference of the feature values between these two sets using a pre-trained facial feature classifier. We took the target feature as the denominator and divided all other feature differences to obtain the corresponding changing ratios based on the target feature. 

We show the top-k difference within the 40 features, which the measurement indicates the $k$ most changed features during transformation.

We randomly selected three features, ``Male'', ``Bald'' and ``Black Hair'' as the evaluated features.
Table \ref{tab:Male quantitative} shows the decrease in the disturbance of  ``Big Nose'' and ``Young'' features. 
Although the ratios of ``Wearing Lipstick'' and ``Heavy Makeup'' increase, we argue that this is the related factor during gender transformation and does not harm gesture preservation.
For the ``Bald'' transformation, as Table \ref{tab:Bald quantitative} shows, three unrelated features (e.g., ``Young'', ``Male'' and ``Chubby'') are successfully suppressed by our approach.
The evaluation of Table \ref{tab:Black_Hair quantitative} for ``Black Hair'' transformations shows that our approach successfully decreases the change of ``Bushy Eyebrows'', ``Bag Lips'' and ``Young'' which should not be involved in this transformation. 
\begin{table*}[!t]
  \centering
  \caption{Relative change ratio of the top-k features when the target feature being edited is ``Male''. A lower value indicates better preservation of original images.}
  \begin{tabular}{|c||c||c||c|}
    \hline
    \makecell{Top K features} 
    & Baseline 
    & Ours 
    & \makecell{Ours Value \\ (for Baseline top K Features)} \\
    \hline
    1st &  Male-1 & Male-1 & Male-1 \\
    \hline
    2nd & Wearing Lipstick-0.80 & Wearing Lipstick-0.83 & Wearing Lipstick-0.83 \\
    \hline
    3rd & Big Nose-0.55 & Heavy Makeup-0.58 & $\downarrow$ Big Nose-0.39 \\
    \hline
    4th & Young-0.54 & Wearing Earrings-0.45 & $\downarrow$ Young-0.41\\
    \hline
    5th & Heavy Makeup-0.53 & Young-0.41 & Heavy Makeup-0.58\\
    \hline
  \end{tabular}
  \label{tab:Male quantitative}
\end{table*}
\begin{table*}[!t]
  \centering
  \caption{Relative change ratio of the top-k features when the target feature being edited is ``Bald''. A lower value indicates better preservation of original images.}
  \begin{tabular}{|c||c||c||c|}
    \hline
    \makecell{Top K features} 
    & Baseline 
    & Ours 
    & \makecell{Ours Value \\ (for Baseline top K Features)} \\
    \hline
    1st &  Bald-1 & Bald-1 & Bald-1\\
    \hline
    2nd & Straight Hair-0.55 & Straight Hair-0.62 & Straight Hair-0.62 \\ 
    \hline
    3rd & Young-0.47 & Brown Hair-0.42 & $\downarrow$ Young-0.32 \\
    \hline
    4th & Male-0.41 & Black Hair-0.35 & $\downarrow$ Male-0.23 \\
    \hline
    5th & Chubby-0.40 & Young-0.32 & $\downarrow$ Chubby-0.31\\
    \hline
  \end{tabular}
  \label{tab:Bald quantitative}
\end{table*}
\begin{table*}[!t]
  \centering
  \caption{Relative change ratio of the top-k features when the target feature being edited is ``Black Hair''. A lower value indicates better preservation of original images.}
  \begin{tabular}{|c||c||c||c|}
    \hline
    \makecell{Top K features} 
    & Baseline 
    & Ours 
    & \makecell{Ours Value \\ (for Baseline top K Features)} \\
    \hline
    1st & Black Hair-1 & Black Hair-1 & Black Hair-1 \\
    \hline
    2nd & Bushy Eyebrows-0.46 &  Brown Hair-0.38 &$\downarrow$ Bushy Eyebrows-0.33\\
    \hline
    3rd & Big Lips-0.41 & Bushy Eyebrows-0.33 &$\downarrow$ Big Lips-0.17 \\
    \hline
    4th & Young-0.38 & Straight Hair-0.28 & $\downarrow$Young-0.25  \\
    \hline
    5th & Brown Hair-0.36 &  Young-0.25 & Brown Hair-0.38\\
    \hline
  \end{tabular}
\label{tab:Black_Hair quantitative}
\end{table*}
We argue that these top-k features inevitably hurt the gesture preservation to some extent.  By suppressing them, we preserve the gesture which is beneficial for data augmentation in gesture research.

To further show the advantages of our approach, we collected a set of explicit and direct features that have a direct impact on facial gesture. The results are shown in Table \ref{gesture feature table}. 
For all three transformations, the disturbance of ``Smiling'' and ``Mouth Slightly Open'' is significantly decreased, with some being reduced to as low as one half or even one third, compared to the corresponding value in the baseline.
We consider the ``Narrow Eyes'' an explicit factor which effect the gesture and expression, since it can make a face slightly angry.
Although for ``Narrow Eyes'' in ``Bald'' transformation, the ratio slightly increase, for most cases (e.g., ``Male'' and ``Black Hair'), the ratio has been significantly decreased.

We then used DeepFace~\cite{serengil2021lightface} to detect the emotion changes during transformation.
We used the same two opposite sets of images mentioned above (e.g., non-exist and must exist for a target feature transformation), we calculated the mean difference of the emotion values between two sets as presented in Table \ref{emotion table}.
Our approach suppresses the changes of all seven emotions among all three transformations.
\begin{table*}[!t]
\caption{The explicit expression and gesture features changing ratio based on target features.A lower value indicates better performance.}
\centering
\begin{tabularx}{\textwidth}{|c||Y Y||Y Y||Y Y|}
\hline
\multirow{2}{*}{\textbf{Gesture Features}} 
& \multicolumn{2}{c||}{\textbf{Male}} 
& \multicolumn{2}{c||}{\textbf{Bald}} 
& \multicolumn{2}{c|}{\textbf{Black Hair}} \\
\cline{2-7}
& baseline & ours & baseline & ours & baseline & ours \\
\hline
Smiling & 0.18 & 0.12 & 0.32 & 0.13 & 0.17 & 0.05 \\
Mouth Slightly Open & 0.29 & 0.20 & 0.30 & 0.15 & 0.29 & 0.07\\
Narrow Eyes & 0.15 & 0.10 & 0.30 & 0.31 & 0.12 & 0.08 \\
\hline
\end{tabularx}

\label{gesture feature table}
\end{table*}

\begin{table*}[!t]
\caption{The emotion disturbance during transformation. A lower value indicates better performance.}
\centering
\begin{tabularx}{\textwidth}{|c||Y Y||Y Y||Y Y|}
\hline
\multirow{2}{*}{\textbf{Emotion}} 
& \multicolumn{2}{c||}{\textbf{Male}} 
& \multicolumn{2}{c||}{\textbf{Bald}} 
& \multicolumn{2}{c|}{\textbf{Black Hair}} \\
\cline{2-7}
& baseline & ours & baseline & ours & baseline & ours \\
\hline
Angry & 11.17 & 8.66 & 16.85 & 14.33 & 12.46 & 9.81 \\
Disgust & 0.57 &  0.51 & 0.74 & 0.65 & 0.33 &  0.26\\
Fear &  7.52 & 5.31 & 8.65 & 8.35 & 7.38 & 5.96 \\
Happy & 14.38 & 10.17 & 21.36 & 16.56 & 24.62 & 12.47 \\
Sad & 20.57 & 16.02 & 28.07 & 24.70 & 25.69 & 19.14 \\
Surprise & 0.99 & 0.81 & 1.39 & 1.04 & 1.29 & 0.87 \\
Neutral & 15.44 & 13.58 &  24.19 & 22.03 & 21.17 & 14.95 \\
\hline
\end{tabularx}

\label{emotion table}
\end{table*}

The quantitative evaluation shows the effectiveness of our proposed HFLD loss. It can reduce the changing ratio for the unrelated features and, thus, performs better disentangled transformation and retaining the gestures.

\subsection{Qualitative Analysis of the Baseline and our model}
The qualitative analysis is consistent with the quantitative comparison.
In ``Female'' transformation, the ``Smiling'' feature always disturbs the transformation in the baseline model. 
After adding HFLD loss during training, the result is more disentangled and the ``expression disturbing'' issue has been reduced significantly.

\begin{figure*}[!t]
  \centering
   \includegraphics[width=0.9\linewidth]{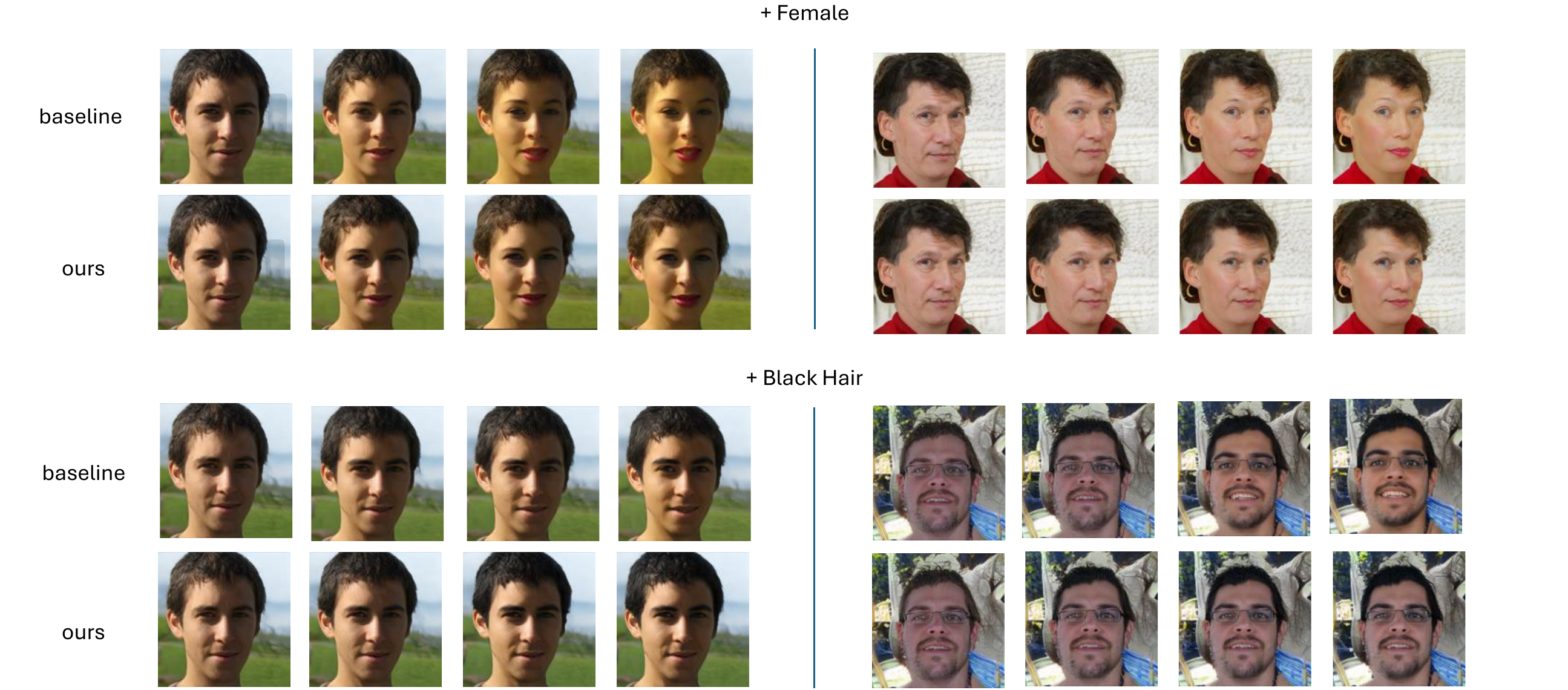}
   \caption{The comparison between baseline (w/o HFLD loss) and ours (w/ HFLD loss). Four sets of transformation trajectory are shown, where, in each set, the first row and second row are the results from baseline and our model, respectively. The original image used for transformation is the leftmost image in each set.}
   \label{C2base}
\end{figure*}

\figref{C2base} illustrates that for the ``Black Hair'' transformation, the baseline not only changes the hair color but also changes the expression. 
For both examples, the baseline makes the mouth slightly open and makes the person happier whereas ours can preserve the facial gesture and expression very well.

\begin{figure*}[!t]
  \centering
   \includegraphics[width=0.82\linewidth]{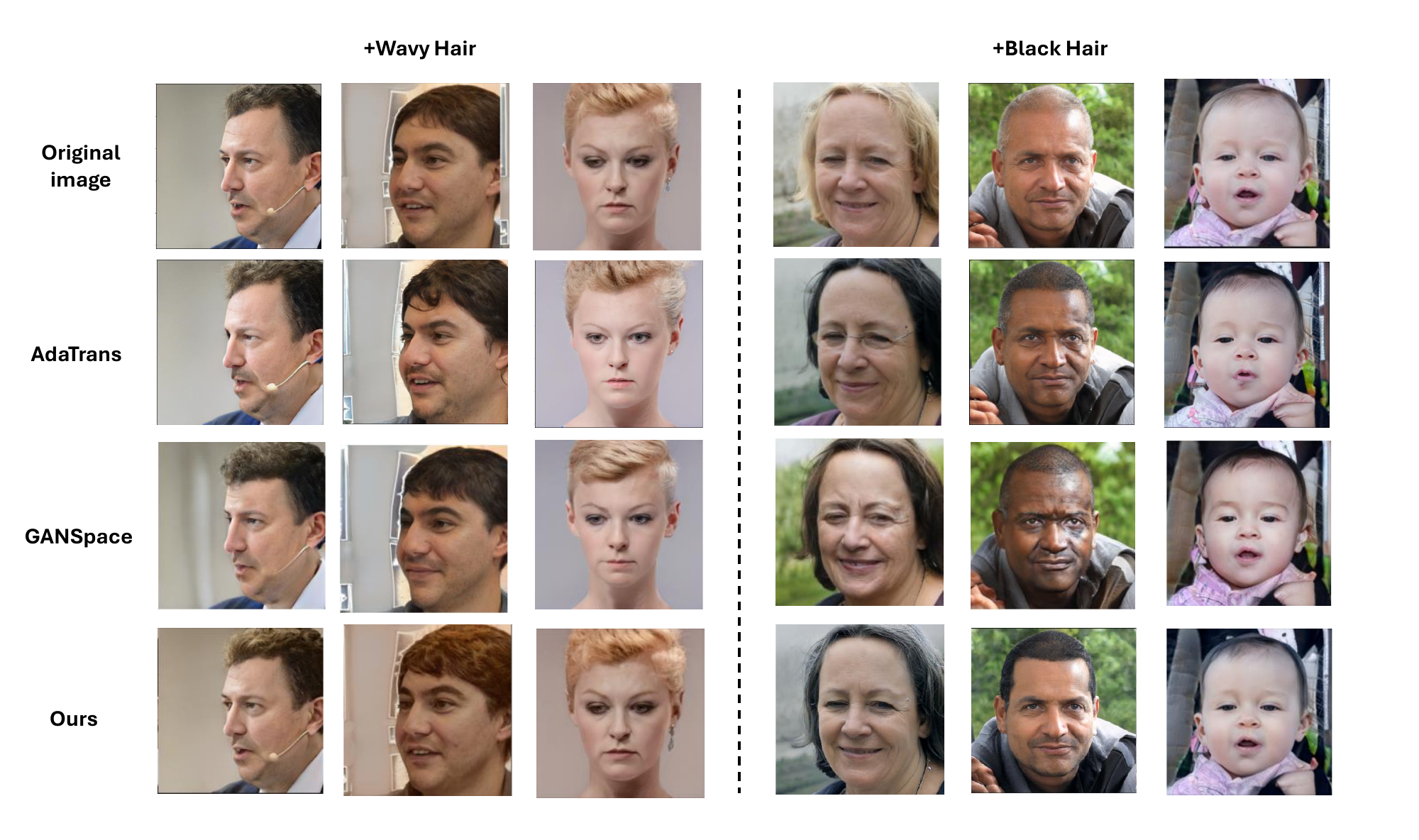}
   \caption{The comparison among AdaTrans, GANSpace and ours. All images in the first row are original images. The second, third and last row show the results from AdaTrans, GANSpace and our model, respectively.}
   \label{C2AdaTrans}
\end{figure*}

\begin{figure*}[!t]
  \centering
   \includegraphics[width=0.65\linewidth]{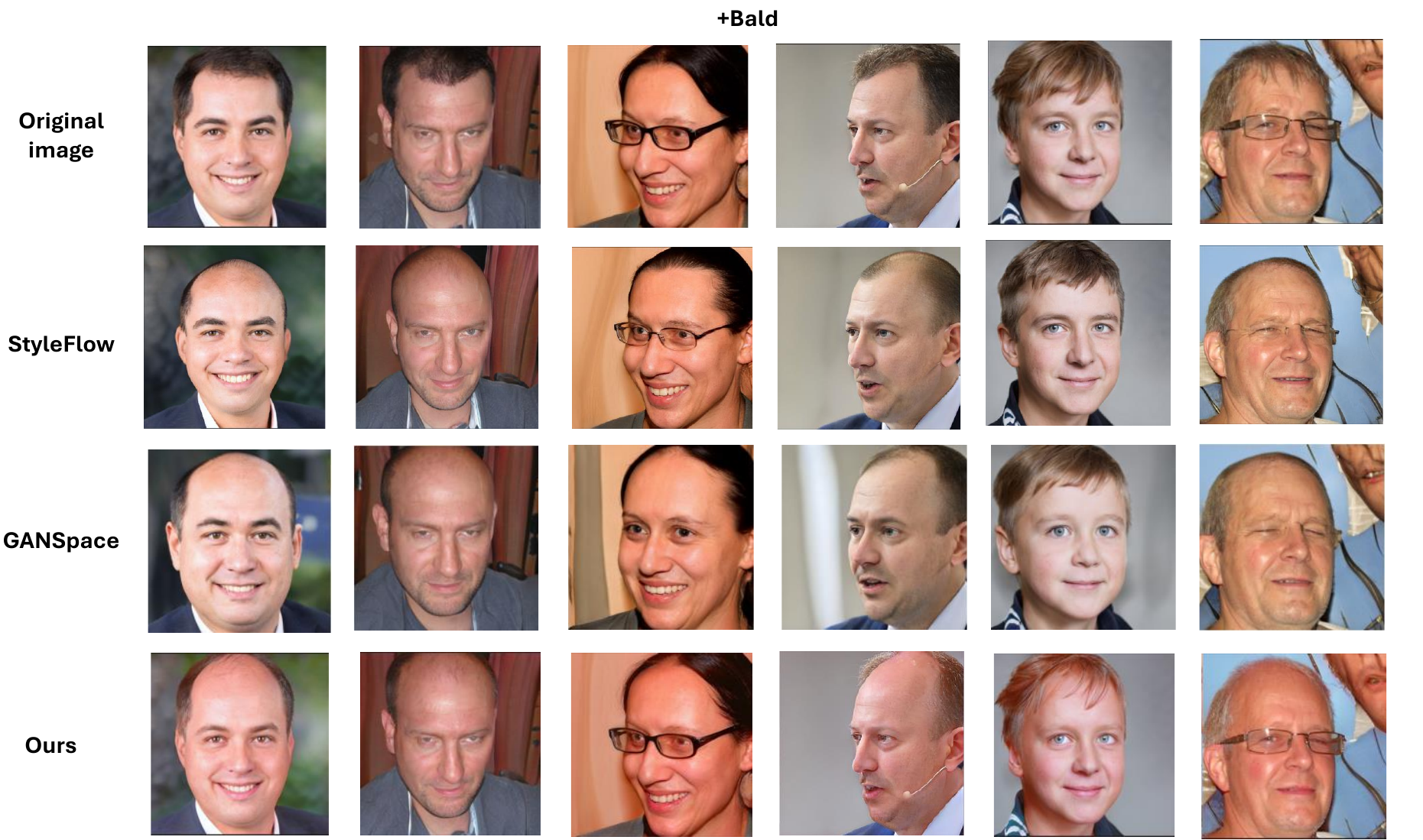}
   \caption{The comparison among StyleFlow, GANSpace and ours. All images in the first row are original images. The second, third and last row show the results from StyleFlow, GANSpace and our model, respectively.}
   \label{C2StyleFlow}
\end{figure*}

\subsection{Qualitative analysis of state-of-the-art models and our model}
To compare and align with other state-of-the-art models, we trained the matrix $T$ 
using a pre-trained StyleGAN2 model with a resolution of $1024 \times 1024$. The weights of the model is the official weights from Nvidia~\cite{karras2019style}.
As shown in \figref{C2AdaTrans}, we used ``Wavy Hair'' and ``Black Hair'' to make an comparison with AdaTrans~\cite{huang2023adaptive} and GANSpace~\cite{harkonen2020ganspace}.

For ``Wavy Hair'' transformation, we observed that the performance of AdaTrans decreases when the head orientation is extreme.
Specifically, the first example makes eyes smaller which hurts the accuracy for expression recognition. 
The third example also changes the eye state which makes the face convey an inaccurate expression.
Similar issues are also observed in GANSpace, the third example changes the mouth and eye state, while ours keeps the expression and gesture better.
In ``Black Hair'' transformation, the expressions are not preserved well in both AdaTrans and GANSpace. 
The first column and third column examples transformed by AdaTrans change the mouth state.
And GANSpace shows poor preservation of skin color feature in the second column, and the mouth is slightly open in the third column example.
In our approach, we significantly reduce these issues.

Since StyleFlow has no options for hair changing and other complex transformations, we picked the ``Bald'' transformation for comparison, seen in \figref{C2StyleFlow}. 
Comparing to StyleFlow, our approach not only preserves the face shape better (c.f. the first and third examples), but also preserves the details such as glasses (c.f. the third and sixth examples) and hook headphones (c.f. the fourth example).
We can also observe that StyleFlow adds a slight smile to the original images, which is especially obvious in the first, second, fourth, and sixth examples where ours keeps the expressions and facial gestures as in the original images.
Comparing to GANSpace, our approach keeps the face shape better (c.f. the first, third, fourth and fifth example) and preserves more details such as glasses and hook headphones as mentioned above.


\section{CONCLUSIONS AND FUTURE WORKS}
\label{sec: conlusion}
\subsection{Conclusions}
We proposed a new way of data augmenting for facial expression and gesture research.
By adding our proposed loss addition, referred to as Human Face Landmark Detection (HFLD) loss, to the baseline model (Enjoy your Editing~\cite{zhuang2021enjoy}), we show significant improvement in terms of preserving the gesture and expression for human face image generation.
With this advantage, we offer an effective option for data augmentation for human face gesture and expression detection.
Given an image, we can sample various different human face images with different appearances such as the gender, hair color, or hair type, while fixing gestures and expressions. 
Our approach significantly saves the labeling cost and provides a reliable data expansion approach.

\subsection{Future Works}
We argue that the addition of HFLD loss can be applied easily to many latent space editing approaches. 
We would expect all GAN approaches, such as AdaTrans~\cite{huang2023adaptive} and MaskFaceGAN~\cite{pernuvs2023maskfacegan}, to benefit from our approach to achieve better expression preservation. 
For editing approaches using diffusion models we expect similar results. 
For example, ChatFace~\cite{yue2023chatface} allows users to edit images using natural language prompts and trains an MLP network to predict the semantic offset, guided by a set of losses.
As future work we seek to validate our assumptions here.
%


\section*{ETHICAL IMPACT STATEMENT}
The use and generation of human faces will always have ethical impacts both positive and negative. On the positive side we are reducing the need to collect and use images of real people (potentially without their concent), whilst on the negative side we are providing a tool which could be used to allow people to generate images of people which could be used for unethical purposes. In this work we use pre-trained models for face generation (StyleGAN and StyleGAN2) thus not requiring the collection of further real-world images. As our approach only allows the retention of gestures between images this will have limited benefit for those wishing to produce unethical images.



{\small
\bibliographystyle{ieee}
\bibliography{sample_FG2025}
}

\end{document}